\documentclass[letterpaper, 10 pt, conference]{ieeeconf}  

\IEEEoverridecommandlockouts                              

\usepackage{graphics} 
\usepackage{epsfig} 
\usepackage{mathptmx} 
\usepackage{times} 
\usepackage{amsmath} 
\usepackage{amssymb}  

\usepackage[switch]{lineno}
\usepackage{booktabs}
\usepackage{threeparttable}
\usepackage{romannum}
\usepackage{soul}

\usepackage{amsthm,amsmath,amssymb}
\usepackage{mathrsfs}

\usepackage{color, xcolor}
\usepackage{here}
\usepackage{graphicx}
\usepackage{subfigure}
\usepackage{cite}
\usepackage{bm}
\usepackage{float}
\usepackage{mathtools}
\usepackage{amsmath,amsfonts}
\usepackage{algorithm}
\newcommand{\nonl}{\renewcommand{\nl}{\let\nl\oldnl}}%

\title{\LARGE \bf
Jumping Control for a Quadrupedal Wheeled-Legged Robot via NMPC and DE Optimization
}

\author{ Xuanqi Zeng$^{1}$, Lingwei Zhang$^{1}$, Linzhu Yue$^{1}$, Zhitao Song$^{1}$, Hongbo Zhang$^{1}$, \\
Tianlin Zhang$^{1}$, and Yun-Hui Liu$^{1}$ 
\thanks{$^{1}$ X. Q. Zeng, L. W. Zhang, L. Z. Yue, Z. T. Song, H. B. Zhang, T. L. Zhang, and Y.-H. Liu are with the Department of Mechanical and Automation Engineering, The Chinese University of Hong Kong.
        {\tt\small xqzeng@mae.cuhk.edu.hk}}%
\thanks{This work is supported by the InnoHK Clusters of the Hong Kong SAR Government via the Hong Kong Centre for Logistics Robotics, and the CUHK T Stone Robotics Institute.}
}

\begin{document}
\maketitle
\thispagestyle{empty}
\pagestyle{empty}

\begin{abstract}
Quadrupedal wheeled-legged robots combine the advantages of legged and wheeled locomotion to achieve superior mobility, but executing dynamic jumps remains a significant challenge due to the additional degrees of freedom introduced by wheeled legs. This paper develops a mini-sized wheeled-legged robot for agile motion and presents a novel motion control framework that integrates the Nonlinear Model Predictive Control (NMPC) for locomotion and the Differential Evolution (DE) based trajectory optimization for jumping in quadrupedal wheeled-legged robots. The proposed controller utilizes wheel motion and locomotion to enhance jumping performance, achieving versatile maneuvers such as vertical jumping, forward jumping, and backflips. Extensive simulations and real-world experiments validate the effectiveness of the framework, demonstrating a forward jump over a 0.12 m obstacle and a vertical jump reaching 0.5 m.


\end{abstract}


\section{INTRODUCTION}
Recent advancements in robotics have led to the development of hybrid robotic platforms that combine different locomotion modes to achieve superior mobility and adaptability in complex environments \cite{eth_wheel_leg_sr, eth_wheel_leg_1, bit_1}. One such example is the quadrupedal wheeled-legged robot, which integrates both legs and wheels to navigate diverse terrains efficiently. These robots are capable of executing complex movements such as walking, rolling, and jumping, each of which presents unique challenges in motion control.

Jumping, in particular, is a critical capability that enhances the robot's versatility, allowing it to overcome obstacles, leap across gaps, and adapt to dynamic environments.  However, achieving dynamic jumping on quadrupedal wheeled-legged robots remains challenging and under-explored. Therefore, how to deal with the extra DoFs of wheeled-legged robots for aiding the driving-jumping motion is an interesting and challenging research topic.

The primary objective of this paper is to investigate and propose a comprehensive motion control strategy for jumping in quadrupedal wheeled-legged robots. The study aims to enhance the quadrupedal wheeled-legged robot's jumping ability by combining locomotion control and jumping trajectory optimization. With the proposed controller, the robot leverages the benefits of wheeled motion to have a great performance in versatile jumping motions, such as vertical jumping, forward jumping, and backflip. To our knowledge, no prior research work has demonstrated hardware forward jumping on a quadrupedal wheeled-legged quadruped with a model-based planning and control pipeline.

\begin{figure}[t]
  \centering
\subfigure[The forward jumping over an obstacle]{
    \includegraphics[width=0.47\textwidth]{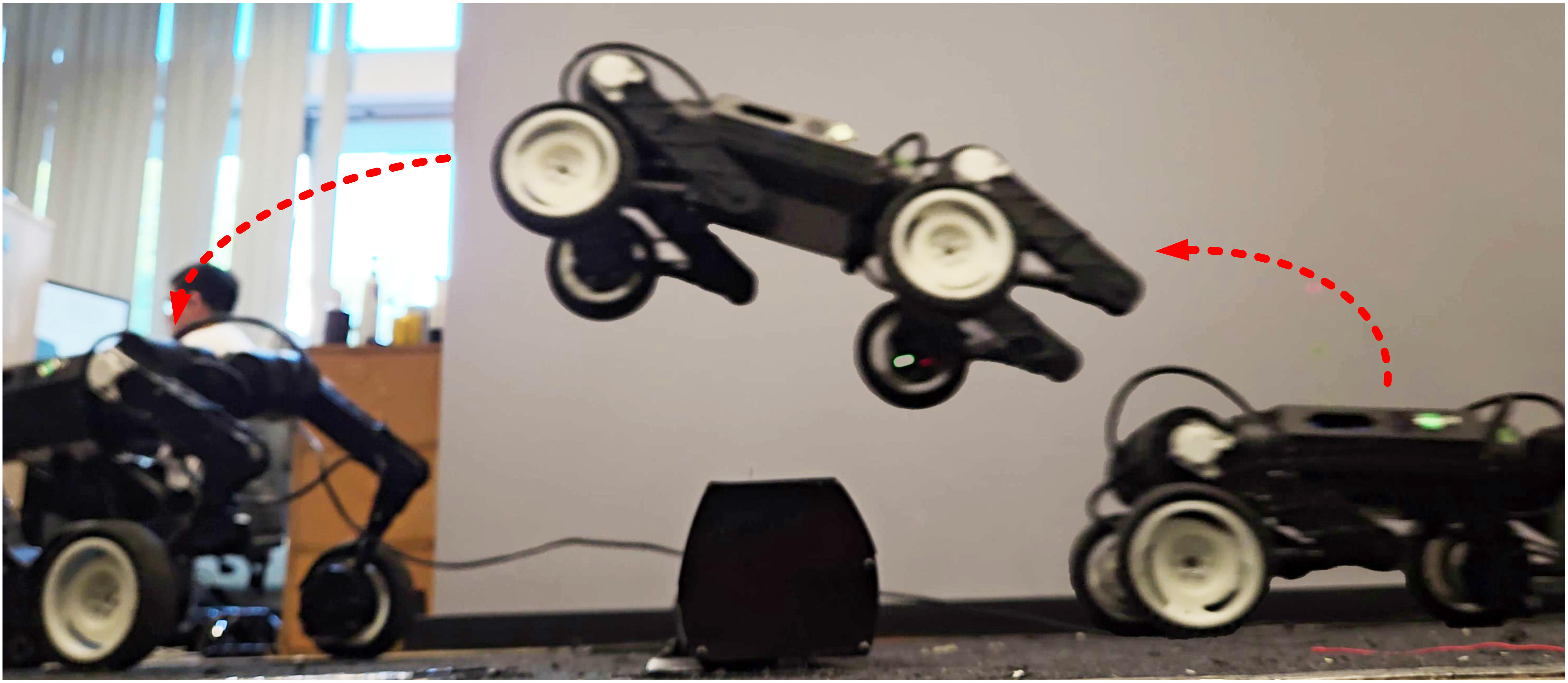}
    \label{foward_jump_1}
    }
\subfigure[The vertical jumping]{
    \includegraphics[width=0.47\textwidth]{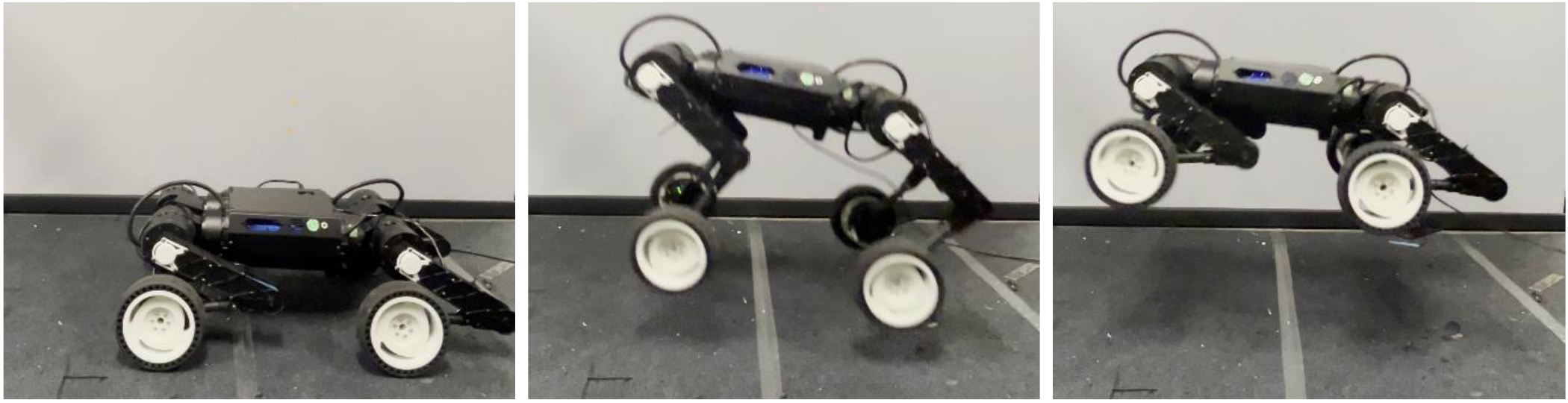}
    \label{vertical_jump_1}
    }
\vspace{-0.5cm}
  \caption{The jumping motion on our quadrupedal wheeled-legged robot: (a) The robot jumping over a 0.12 m obstacle; (b) The robot achieves a vertical jump with a maximum height of 0.5 m.}
  \label{forward_vertical_jumping} 
\vspace{-0.5cm}
\end{figure}

\subsection{Related Work} \label{related_work}
\paragraph{Wheeled-legged Robots}
Wheeled-legged robots can be classified into two categories based on the actuation method of their wheels: (a) passive wheels and (b) active wheels. Robots equipped with active wheels offer greater efficiency and flexibility, making them suitable for real-world applications. However, achieving agile locomotion in wheeled-legged robots remains a challenging research problem. While robots such as Handle \cite{handle}, Ascento \cite{ascento}, Centauro \cite{centauro}, and ANYmal \cite{anymal} demonstrate advanced capabilities but are primarily confined to research and development environments. On the other hand, legged robots with passive wheels benefit from lighter wheels and can be controlled in a manner similar to traditional legged robots, potentially enhancing mobility efficiency. However, the works related to this kind of robot are few \cite {skate_tmech, skate_eth, skate_icra}. Their limitations are obvious: they can only skate on pretty even ground, and passive sliding motion is hard to control and deteriorates their locomotion performance, which makes them far beyond application.

\paragraph{Model Predictive Control} Model-based control strategies, especially Model Predictive Control (MPC), have gained considerable traction in the field of quadruped robotics owing to their exceptional capabilities. For instance, the MIT Cheetah 3 and Mini-Cheetah leverage convex MPC to enable stable and versatile locomotion, supporting multiple gaits such as trotting, bounding, and galloping \cite{Di_01}, \cite{Kim_02}. Nonetheless, these methods often depend on the simplification of treating the system as a Single Rigid Body Dynamics (SRBD) to strike a balance between computational efficiency and model fidelity. To address this limitation and improve control accuracy, recent advancements have focused on incorporating more sophisticated dynamic models, with Nonlinear Model Predictive Control (NMPC) being applied to enhance the performance of legged robotic systems \cite{Neunert_03, Meduri_04, qiayuan_01}.

\paragraph{Jumping Trajectory Optimization}
Just like their animal counterparts, jumping is also a vital ability for enabling legged robots to traverse challenging terrains. Hence, some related works concentrate on trajectory optimization for legged robots to achieve a jumping motion. For example, Chignoli et al.\cite{hierarchical_jump} present a hierarchical planning framework for online omnidirectional jumping; Nguyen et al.\cite{MIT_jump} propose an efficient trajectory optimization that enables MIT Cheetah 3 to jump onto a 0.76 m high. A two-legged planar robot executes a jump to a 0.2 m high platform by the mixed-integer convex program in the paper of Ding et al.\cite{Ding_jump}. However, the research related to the jumping motion
of quadruped wheeled-legged robots is little. There is a paper where a quadruped wheeled-legged robot achieves driving-jumping in the simulation but fails in real-world experiments due to hardware issues \cite{sim_jump}. 

\begin{center}
\vspace{-0.3cm}
\begin{figure}[htb]
\centering
\includegraphics[width=2.2in]{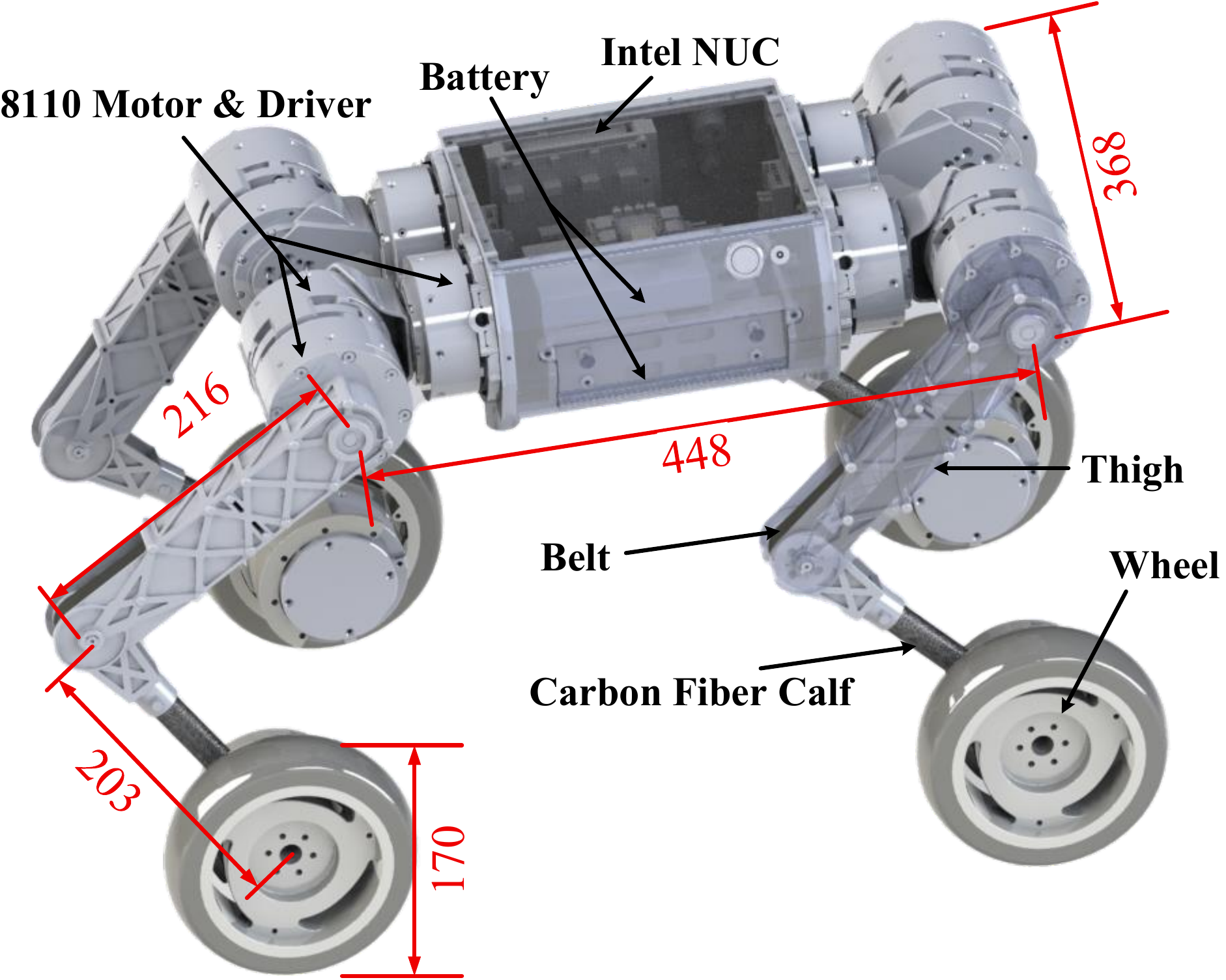}
\vspace{-0.3cm}
\caption{The hardware design of our mini-size quadrupedal wheeled-legged robot (unit: mm).}
\label{hardware_design}
\end{figure}
\vspace{-0.3cm}
\end{center}

\subsection{Contribution} 
The summary of our main contributions is as follows: 
\begin{itemize}
\item {We design a quadrupedal wheeled-legged robot as shown in Fig. \ref{hardware_design} and propose a jumping control framework with two components: (i) an NMPC locomotion controller to accelerate the robot to the desired velocity in preparation for the jump, and (ii) the jumping trajectory optimization by Differential Evolution (DE) algorithm generates optimal inputs for the system in the take-off phase to ensure a successful jumping motion. The control framework is presented in Fig. \ref{control_framework}.}
\item {We demonstrate the effectiveness of our controller through a series of simulations and physical experiments conducted on our robot. These tests confirm that the proposed control framework enables stable locomotion and reliable jumping capabilities. In hardware trials, the robot successfully performs a forward jump to clear a 0.12 m obstacle and achieves a vertical jump reaching a height of 0.5 m, showcasing its dynamic performance.}
\end{itemize}
\begin{center}
\vspace{-0.3cm}
\begin{figure}[htb]
\centering
\includegraphics[width=3.3in]{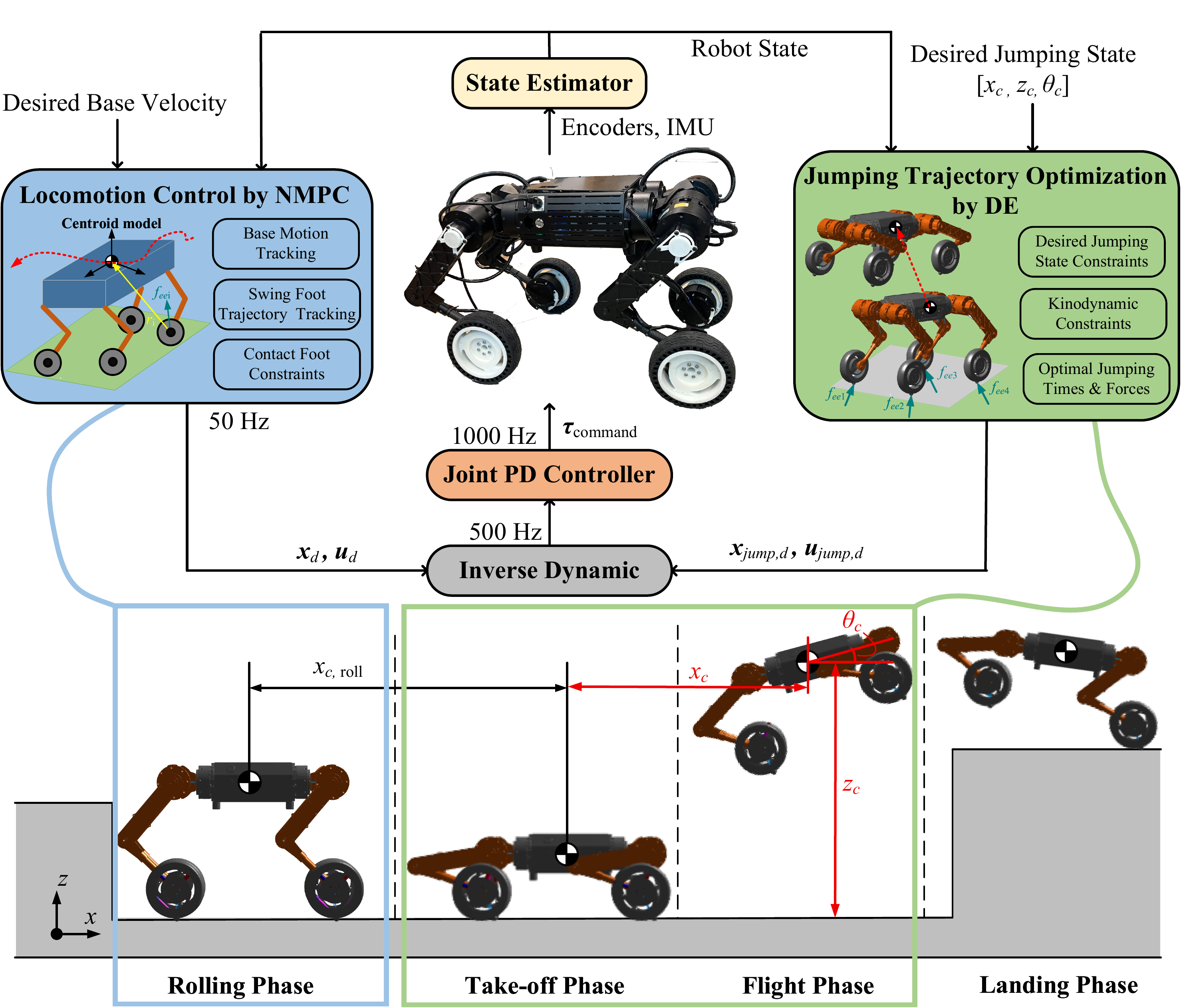}
\vspace{-0.2cm}
\caption{Overview of the control framework. The locomotion of our wheeled-legged robot is implemented by Nonlinear Model Predictive Control (NMPC), which generates the required forward velocity before the jump. The Differential Evolution (DE) algorithm is activated only during the take-off phase to solve the jumping trajectory optimization problem. The $d$ subscript indicates the desired value. The term $x_{c, \textnormal{roll}}$ represents the displacement of the Center of Mass (CoM) in the $x$-direction before the flight phase. The desired jumping state in the $x$-$z$ plane, denoted as $[x_c, z_c, \theta_c]$, corresponds to the desired jumping distance, height, and base rotation angle, respectively.}
\label{control_framework}
\end{figure}
\vspace{-0.3cm}
\end{center}

\section{Locomotion Control of Quadrupedal Wheeled-legged Robot By NMPC} \label{NMP_formulation}

In this section, we introduce the NMPC framework designed for the wheeled-legged robot. The locomotion controller plays a key role in generating the initial velocity required for the robot prior to executing a jump, thereby enhancing its overall jumping performance.

\subsection{Centroidal Dynamics Model}
In the application of MPC to quadruped robots, the robot's body is often simplified as an SRBD to enhance computational efficiency. The SRBD model approximates the legged robot's dynamics as a rigid body with massless legs, significantly reducing complexity. However, for quadrupedal wheeled-legged robots, the mass and inertial properties of the legs contribute substantially to the overall system dynamics and cannot be ignored. To address this, our locomotion control approach for wheeled-legged robots is grounded in the centroidal model. This model captures the floating base dynamics of the robot, which are defined as follows:
\begin{equation}
\begin{aligned}
M_u \bm {\ddot q} + h_u(\bm q, \dot{\bm q}) = J_{c,u} ^T\bm F
\end{aligned}
\label{eq:centroidal_model_dynamic}
\end{equation}
where $\bm q \in \mathbb{R}^{6 + n_j}$ is the generalized coordinates. $M_u \in \mathbb{R}^{6 \times (6 + n_j)}$ denotes the un-actuated part of the generalized mass and $h_u \in \mathbb{R}^{6}$ represents the un-actuated part of the vector of Coriolis, centrifugal and gravitational terms. $J_{c,u} \in \mathbb{R}^{n_c \times 6}$ presents the un-actuated contact Jacobians, and $\bm F$ is the vector of Ground Reaction Forces (GRFs).

After applying the transformation on Equation (\ref{eq:centroidal_model_dynamic}), the centroidal dynamics is given by:
\begin{equation}
\dot {\bm{h}}_{com} =\begin{bmatrix} \sum\limits_{i = 1}^{n_c}\bm{f}_{eei} + m \bm{g} \\ \sum\limits_{i = 1}^{n_c}\bm{r}_{i} \times \bm{f}_{eei}\end{bmatrix}
\label{eq:com_centroidal}
\end{equation}
where ${\bm{h}}_{com}=(\bm{P}_{com}, \bm{L}_{com})$ $\in \mathbb{R}^{6}$ presents the centroidal momentum including the linear momentum $\bm{P}_{com}$ and the angular momentum $\bm{L}_{com}$. $\bm{r}_i$ is the position vector from the CoM to the contact point $ci$. ${\bm f_{eei}} \in \mathbb{R}^{3}$ denotes the GRFs for the corresponding limbs.  $\bm g$ is the gravitational acceleration term.

\begin{figure}[htb]
    \centering
    \vspace{-0.3cm}
    \includegraphics[height=0.25\textwidth]{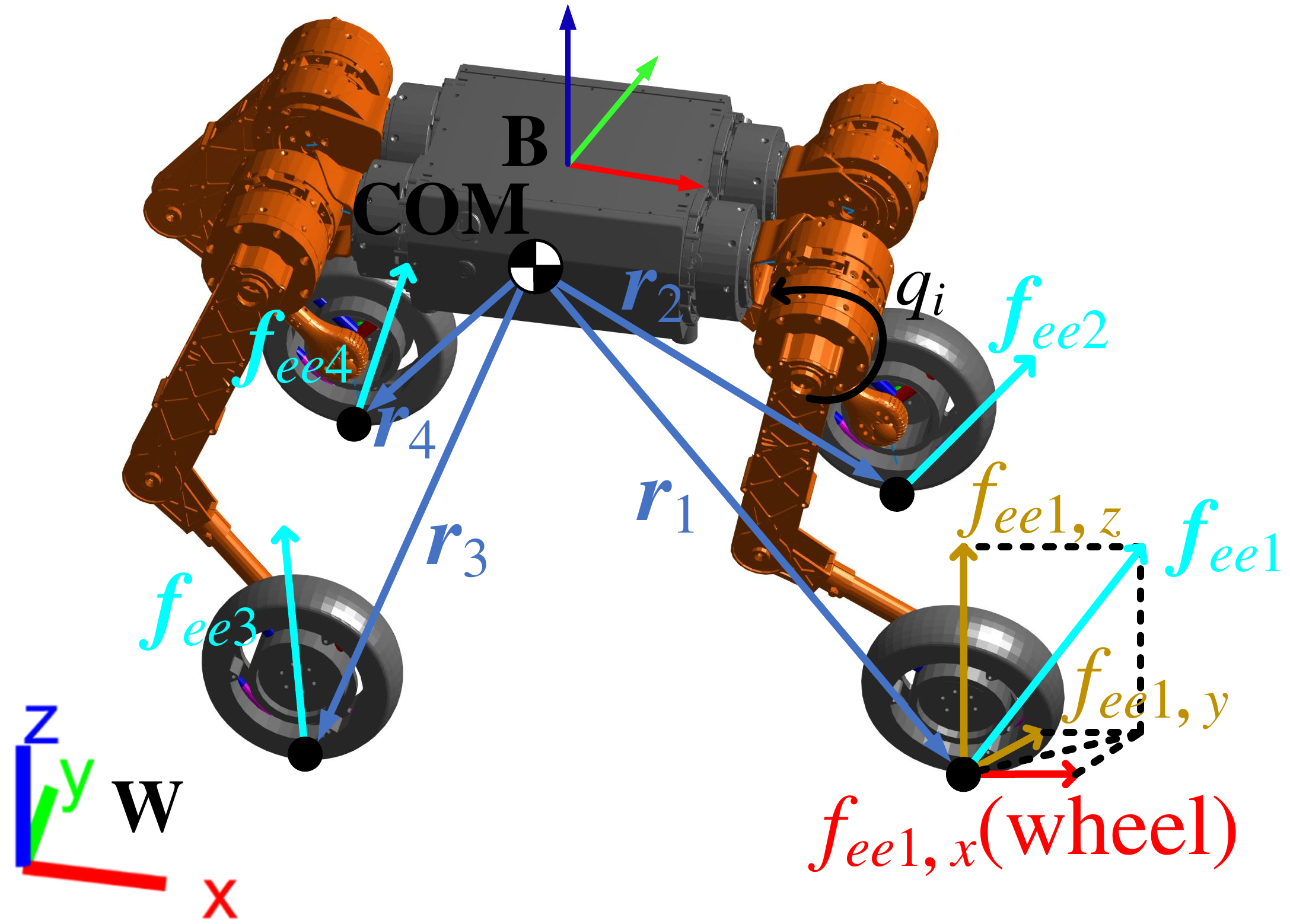}
    \vspace{-0.3cm}
    \caption{The model of centroidal dynamic on our robot. $f_{ee1, x}$, $f_{ee1, y}$, $f_{ee1, z}$ indicates the GRF $\bm f_{ee1}$ in the $x$, $y$ ,$z$ direction w.r.t Body Frame.}
    \label{centroidal_model}
    \vspace{-0.0cm}
\end{figure}

The centroidal dynamic of our quadrupedal wheeled-legged robot is shown in Fig. \ref{centroidal_model}. The state and the input of the system are defined as follows:
\begin{equation}
\begin{aligned}
{\bm x}=[{\bm \theta}^T ~ {\bm p}^T ~ {\bm \omega}^T ~ {\bm v}^T ~ {\bm q_j}^T]^T
\end{aligned}
\label{eq:state}
\end{equation}
\begin{equation}
\begin{aligned}
{\bm u}=[{\bm f_{ee1}}^T ~ {\bm f_{ee2}}^T ~ {\bm f_{ee3}}^T ~ {\bm f_{ee4}}^T~ \dot{\bm q_j}^T]^T
\end{aligned}
\label{eq:state}
\end{equation}
where ${\bm \theta}, {\bm p}, {\bm \omega}, {\bm v} \in \mathbb{R}^{3}$ represent the Euler angles, positions, angular velocities, and linear velocities of the robot's body in the world frame, respectively. The joint positions and velocities of the robot are denoted by ${\bm q_j}$, $\dot{\bm q}_j \in \mathbb{R}^{n_j}$, respectively.

\subsection{NMPC Problem Formulation}
Locomotion control in legged robots focuses on the crucial objective of controlling the robot's body position and orientation to match desired trajectories. To accomplish this, the NMPC framework solves an optimization problem to compute the optimal GRFs and joint velocities, ensuring the robot achieves its target states based on predicted future dynamics. The nonlinear optimal control problem is formulated as follows:
\begin{subequations}
\begin{align}
\min_{\bm u} \quad \quad &\sum _{k=0} ^{N-1} l({\bm {x_{t+k}}},{\bm { u_{t+k}}})\\
\label{discrete_form_centoridal_dynamic}
{s.t.} \quad \quad  &{\bm {x_{t+k+1}}} =f({\bm x_{t+k}}, {\bm u_{t+k}}) , \quad  k=0,1,...,N-1\\
&{\bm x_t}=\bm x(t)\\
&g_k({\bm {x_{t+k}}},{\bm { u_{t+k}}})=0, \quad k=0,1,...,N\\
&h_k({\bm {x_{t+k}}},{\bm { u_{t+k}}})\geq0, \quad k=0,1,...,N
\end{align}
\end{subequations}
where Equation (\ref{discrete_form_centoridal_dynamic}) represents the discrete time form of the centroidal dynamics. $l(\cdot)$ denotes the stage quadratic cost function, ${\bm x_t}$ is the current state at time step $t$, and ${\bm x_{t+k}}$ corresponds to the predicted state at time step $t+k$ based on the centroidal dynamic. ${N}$ represents the prediction horizon. Additionally, $g_k(\cdot)$ and $h_k(\cdot)$ define the general equality and inequality constraints, respectively.

\subsection{Foot Constraints}\label{section_foot_constraint}
For this floating base system of quadrupedal wheeled-legged robots, the foot contact with the ground can provide inputs for the system to control the desired base motion. The foot constraints in this optimal control framework consist of the swing foot constraint and the contact foot constraint.

The swing foot constraints, analogous to those used in legged robot control, ensure that the foot follows a predefined spline curve to reach the desired foot height as well as guarantee that there is no external force exerted on the foot:
\begin{equation}
\textnormal{Swing foot constraints:} \left\{ \begin{aligned}
& \bm{v_{eei}} \cdot \bm{n} = \bm{v}_{desired}(t)\\
& \bm{f}_{eei} = 0
\end{aligned} \right.
\label{eq:swing_constraint}
\end{equation}
where $\bm{n}$ presents the normal vector of the ground, $\bm{v}_{eei} \in \mathbb{R}^{3}$ denotes the velocity of the corresponding foot, $\bm{v}_{desired}$ is the desired foot velocity in the normal direction.

\begin{center}
\begin{figure}[htb]
\vspace{-0.3cm}
\centering
\includegraphics[width=2.5in]{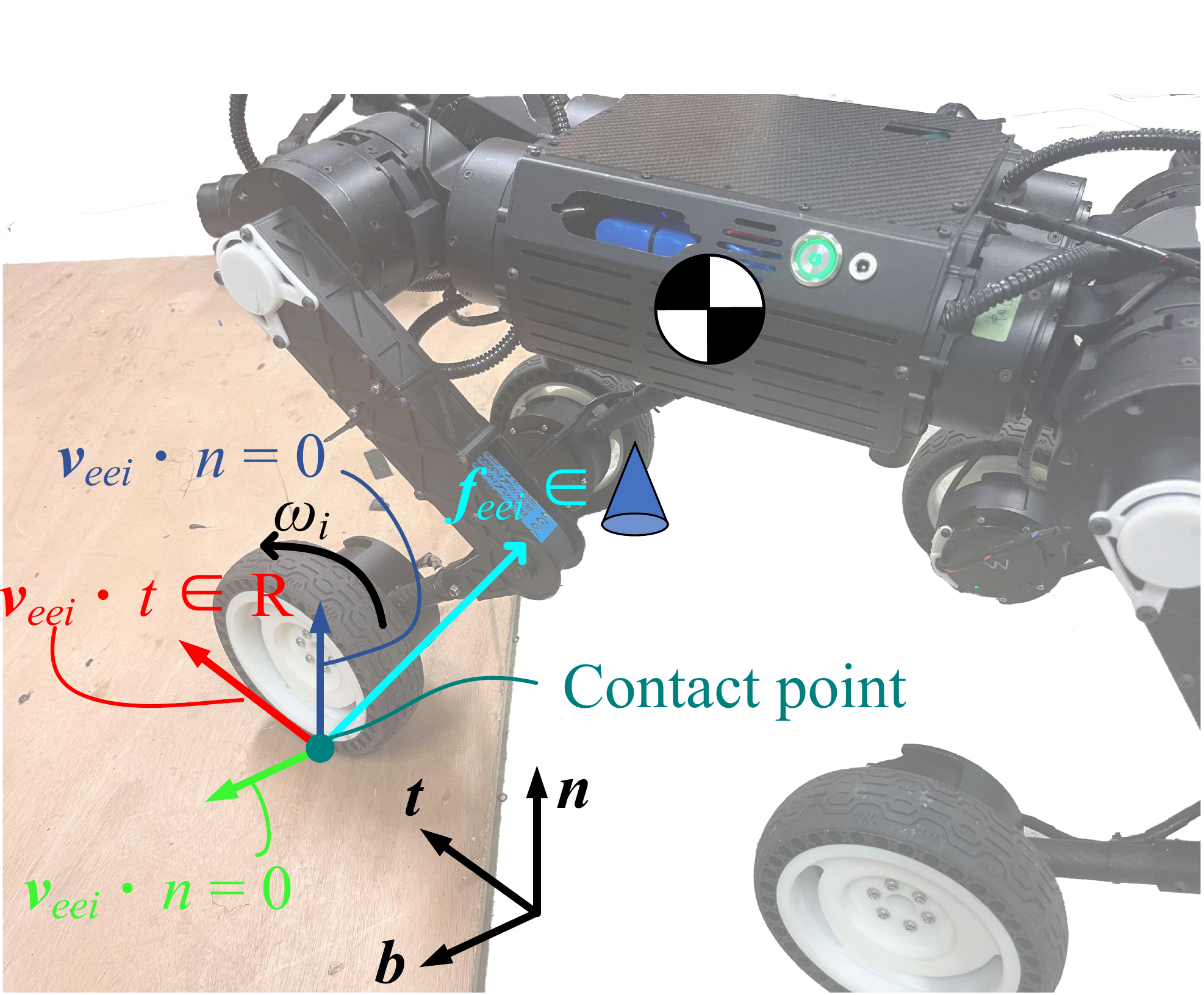}
\vspace{-0.3cm}
\caption{The foot's constraints when the wheel contacts the ground. It includes the velocity constraints of the wheel in the three directions ($\bm t $, $\bm n $, $\bm b $) and the friction cone constraints of foot force $\bm f_{eei}$. $\omega_i$ denotes the rotation speed of the wheel.}
\label{contact_feet_constraints}
\vspace{-0.5cm}
\end{figure}
\end{center}

The contact foot constraints deal with the foot's motion when it comes in contact with the ground. On the one hand, it constrains the foot force within the friction cone so as not to slide between the contact foot and the ground. On the other hand, it requires that the foot velocity must be zero in the lateral and vertical directions. In contrast, the foot velocity along the wheel’s rolling direction ($\bm t$ vector direction in Fig. \ref{contact_feet_constraints}) is free \cite{anymal}, as the wheel can roll without slipping in that direction, which distinguishes wheeled-legged robots from purely legged systems. The details of the contact foot constraints are shown in Fig. \ref{contact_feet_constraints}. The contact foot constraint of this system is presented by the following:
\begin{equation}
\textnormal{Contact foot constraints:} \left\{ \begin{aligned}
& \bm{v_{eei}} \cdot \bm{n} = 0\\
& \bm{v_{eei}} \cdot \bm{b} = 0\\
& \bm{v_{eei}} \cdot \bm{t} \in R \\
& \frac{\sqrt{f_{eei,x}^2+f_{eei,y}^2}}{f_{eei,z}} < \mu
\end{aligned} \right.
\label{eq:contact_contrain}
\end{equation}
where $\bm t$, $\bm n$, and $\bm b := \bm n \times \bm t$ are the unit tangent (wheel rolling direction), surface normal, and binormal of the contact frame, respectively. $\mu$ is the Coulomb friction coefficient.

\subsection{Joint Control}\label{Joint_control}
In the joint-level control of the quadrupedal wheeled-legged robot, we employ a Proportional-Derivative (PD) controller for all joints. More specifically, in the control of wheel joints, since we are more concentrated on the velocity tracking accuracy of the robot, the position gain of the wheel motor is set to 0.

The above NMPC problem is solved by Sequential Quadratic Programming (SQP) in an open source library OCS2 \cite{ocs2}, and it outputs the optimal GRFs as well as joint velocities for the system. In order to obtain the actuated torque for the motors, the leg dynamic of the robot is considered:
\begin{equation}
\begin{aligned}
M_a \bm {\ddot q} + h_a(\bm q, \dot{\bm q}) = \bm{\tau}_j + J_{c,a} ^T\bm F
\end{aligned}
\label{eq:leg_dynamic}
\end{equation}
where $\bm \tau_j \in \mathbb{R}^{n_j}$ denotes joint torque of the robot. $M_a \in \mathbb{R}^{n_j \times n_j}$ presents the actuated part of the generalized mass, and $h_a \in \mathbb{R}^{n_j}$ is the actuated part of the vector of Coriolis, centrifugal, and gravitational terms. $J_{c,a} \in \mathbb{R}^{n_c \times n_j}$ presents the actuated contact Jacobians.

The inverse dynamic is calculated based on Equation (\ref{eq:leg_dynamic}) in real-time by Pinocchio\cite{pinocchio}, and the desired position, velocity, and torque for the corresponding joint can be obtained. These commands are then tracked using a PD controller for the leg motors. For the wheel motors, the control law is defined as follows:
\begin{equation}
\tau_{\textnormal{wheel},i} = K_d \cdot (v_{\textnormal{body,desired}, \bm t}-v_{\textnormal{body,current}, \bm t})+f_{eei, \bm t} \cdot r_{\textnormal{wheel}}
\end{equation}
where $\tau_{\textnormal{wheel}, i}$ is the output torque for the wheel motor in the leg $i$. $v_{\textnormal{body,desired}, \bm t}$ and $v_{\textnormal{body,current}, \bm t}$ indicate the desired and current estimated velocities in the $\bm t$ direction respectively. $f_{eei,\bm t}$ presents the optimized GRFs of the leg $i$ in the $\bm t$ direction. $r_{\textnormal{wheel}}$ is the radius of the wheel.

\section{Jumping Motion Optimization}
\label{Jumping_opt}
As previously discussed, the wheeled-legged robot is capable of executing motions even when its legs are in the contact phase. By leveraging the locomotion controller described in Section \ref{NMP_formulation} to supply the desired initial state, the jumping motion optimization module is able to optimize a more aggressive jumping trajectory for the robot.

\begin{figure}[htb]
    \centering
    \vspace{-0.0cm}
    \includegraphics[height=0.14\textwidth]{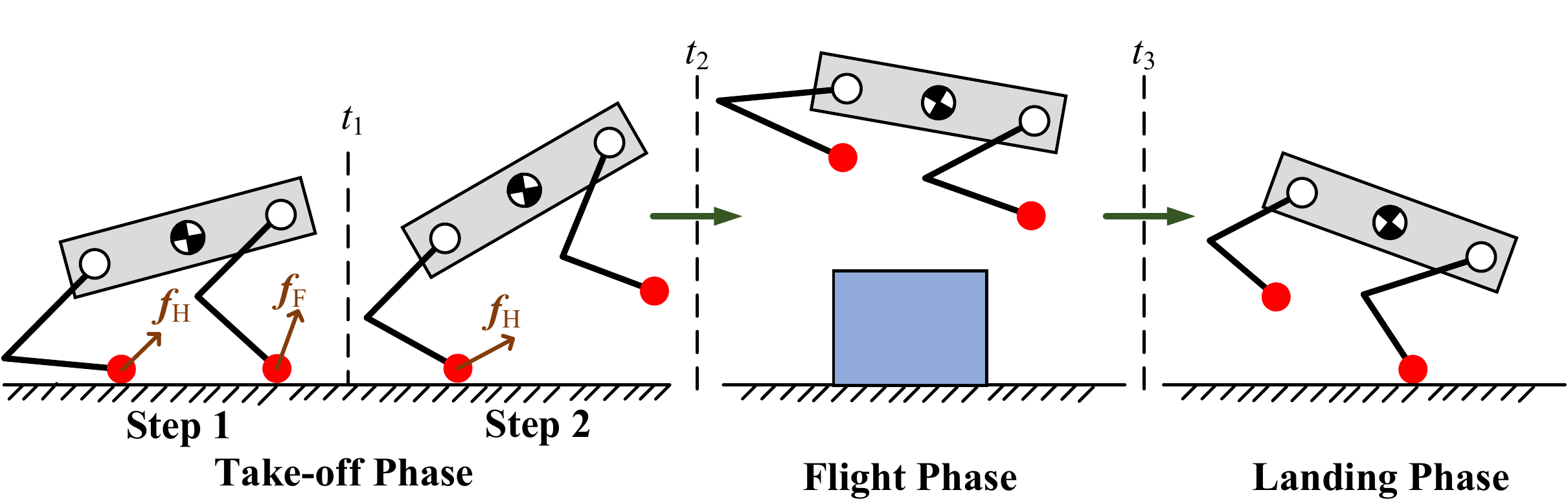}
    \vspace{-0.3cm}
    \caption{Three phases of jumping motion}
    \label{jump_phase}
    \vspace{-0.3cm}
\end{figure}

The DE algorithm is a widely used optimization method known for its high search accuracy, robustness, fast convergence, and ability to find global optimum solutions \cite{DE_review}. These features make it an ideal tool for the trajectory optimizing in legged system. For example, in the work of \cite{yue_omi_jump}, it uses DE algorithm to achieve omnidirectional jumping trajectory planning in legged robots. This section presents the methodology of implementing the jumping trajectory optimization based on the DE algorithm in our quadrupedal wheeled-legged robot. For this method, its optimization is based on the three phases, including the take-off phase, the flight phase, and the landing phase, which is demonstrated in Fig. \ref{jump_phase}.

\subsection{Jumping Trajectory Optimization Formulation} \label{jump_tra_opt}
For the jumping task of the robot, the robot states are defined as the following:
\begin{equation}
{\bm x}_{jump}=[{\bm {p_{com}}}^T ~ ~ {\bm {\Theta}_B}^T ~ ~ {\bm {v_{com}}}^T ~ ~ {\bm {\omega_{B}}}^T]^T
\label{eq:jump_state}
\end{equation}
\begin{equation}
{\bm Q}=[{\bm {q}_{j}}^T ~ ~ {\bm{\dot{q}_{j}}}^T]^T
\label{eq:jump_joint}
\end{equation}
where ${\bm {p_{com}}}$, ${\bm {v_{com}}}$ $\in \mathbb{R}^{3}$ are the position and velocity of the robot body's center of mass (CoM) w.r.t inertial frame. ${\bm {\Theta}_B}$, ${\bm {\omega_{B}}}$ $\in \mathbb{R}^{3}$ represent the Euler angles and angular velocities of the robot's body. 

The objective of the jumping optimization problem is to determine the desired control inputs for the system during the take-off phase, enabling the robot to reach the target position at the end of the jumping motion while satisfying kinodynamic constraints. To enhance motion smoothness, energy consumption is also incorporated into the optimization. The final formulation of the jumping trajectory optimization is as follows:
\begin{subequations}
\begin{align}
\min_{\bm D_{opt}} \quad &\sum _{n=1} ^{L} W_n(10^{n+3} + 10^{n}\sigma_n) +\zeta\\
{s.t.} \quad  &{\bm {x_{jump, k+1}}} ={\bm x_{jump, k}} + \Delta t \bm{\dot{x}_{jump, k}}\\
\label{eq:SRBD}
&{\bm {\dot{x}_{jump, k+1}}} =g({\bm {{x}_{jump, k}}}, {\bm {{u}_{jump, k}}}, {\bm {{Q}_{k}}})\\
&{\bm x_{jump, k}} \in \mathbb{X}, {\bm u_{jump,k}} \in \mathbb{U}, {\bm Q_k} \in \mathbb{Q}, k=0,1,...,N\\
&\bm{x}_{jump}(0)=\bm x_{jump, 0} \\
&\bm{x}_{jump}(N)=\bm{x}_{jump ,N} = \bm{x}_{jump, \textnormal{target}}
\end{align}
\label{jump_opt_formulation}
\end{subequations}
where $D_{opt}$ denotes the optimization variable, which will be introduced in the next part. The term $\sigma_n$ represents the penalty function applied to enforce kinodynamic constraints. $\zeta = \int _0 ^T |\tau(t)\dot{q}(t)|dt$ represents the mechanical energy consumption during the take-off phase. $W_n$ is the penalty coefficient, and $N$ is the total number of the trajectory tracking points. The sets $\mathbb{X}$, $\mathbb{U}$, and $\mathbb{Q}$ define the feasible regions that satisfy the imposed constraints. The terms $\bm x_{jump}(0)$, $\bm x_{jump}(N)$ correspond to the robot's initial and desired jumping states, respectively. Besides, the DE algorithm is hard to online optimize jumping motion based on the centroidal dynamics model. Thus, for reducing computational complexity to achieve online optimization, Equation (\ref{eq:SRBD}) is the SRBD dynamics equation of the robot.

In planar jumping, it is reasonable to simplify the problem by assuming that the force perpendicular to the plane is zero. With this assumption, the GRFs can be decomposed into two components: the total force exerted by the hind legs, $\bm{f}_H$, and the force exerted by the front legs, $\bm{f}_F$. To perform complex jumping maneuvers—such as backflips or jumps onto high platforms—where the contact legs may not leave the ground simultaneously, the take-off phase must be divided into two distinct steps, as shown in Fig. \ref{jump_phase}.

Then, the input of the system is regarded as $\bm {u_{jump}} = [\bm f_H, \bm f_F]$ and assumed as a polynomial curve about time:
\begin{equation}
\bm{u_{jump}} =  \left\{ \begin{aligned}
\bm{a}_1t+\bm{a}_0 \quad & t \in [0, t_1] \\
\bm{\gamma} (\bm{b}_2t^2+\bm{b}_1t+\bm{b}_0) \quad & t \in [t_1, t_2], \bm{\gamma} \in \{0, 1\}\\
0 \quad & t \in [t_2, t_3]
\end{aligned} \right.
\label{eq:u_opt}
\end{equation}
\begin{equation}
\Lambda = [\bm{a}_0, \bm{a}_1, \bm{b}_0, \bm{b}_1, \bm{b}_2] \in \mathbb{R}^N
\label{eq:lambda}
\end{equation}
where $\Lambda$ represents the set of polynomial coefficients that define the curve. The variables $t_1$, $t_2$ denote the end times of steps 1 and 2 in the take-off phase, respectively, while $t_3$ indicates the end of the flight phase. During the time interval $t \in [0, t_1]$,  all legs remain in contact with the ground. In the interval $t \in [t_1, t_2]$, two legs lift off. For $t \in [t_2, t_3]$, the robot enters the flight phase. As shown in Fig. \ref{jump_phase}, when the take-off phase is segmented into two distinct steps, the parameter $\bm{\gamma}$ is set to 1 to indicate that the step 2 (hind-leg push-off) is active. If all four legs leave the ground simultaneously, $\bm{\gamma} = 0$.

\subsection{Optimization Variables}\label{opt_var}
By combining Equations (\ref{eq:u_opt}) and (\ref{eq:lambda}) with the dynamic model of the robot, the optimization problem can be reformulated to optimize the CoM trajectory, jumping phase times, and GRFs. For a plane jump. the robot state can be represented as $\bm{s}_\Omega(t)=[x_c, z_c, \theta_c]$, where $x_c, z_c$ denotes the CoM position in the plane coordinate system and $\theta_c$ represents the body rotation angle. The parameter $\Lambda$ is determined from a sequence of $\bm{s}_\Omega(t)$. In our jumping formulation, the robot's state is evaluated at three specific time points: $\frac{t_1}{2}, t_2, t_3$. Hence, the optimization variables can be given by:
\begin{equation}
\bm{D}_{opt} = [\bm{s}_\Omega (\frac{t_1}{2}), \bm{s}_\Omega (t_2), \bm{s}_\Omega (t_3), t_1, t_2, t_3]
\label{eq:opt}
\end{equation}

\subsection{Kinodynamic Constraints}\label{constraints}
Due to hardware limitations, certain constraints must be integrated into the optimization formulation to ensure that the robot can successfully execute the jump motion in real-world scenarios. These constraints are outlined as follows:
\begin{equation}
\left\{ \begin{aligned}
&\textnormal{Contact Force:} \quad f_{eei,z} > f_{z, min} \\
&\textnormal{Friction Cone:} 
 \quad \frac{\sqrt{f_{eei,x}^2+f_{eei,y}^2}}{f_{eei,z}} < \mu \\
&\textnormal{Joint Angle:} \quad q_{min}<q_{i}<q_{max} \\
&\textnormal{Joint Velocity:} \quad |\dot{q}_{i}|<\dot{q}_{max} \\
&\textnormal{Joint Torque:} \quad |\tau_{i}|<\tau_{max} \\
&\textnormal{Joint Position:} \quad z_B > z_{min} \\
\end{aligned} \right.
\label{eq:constraints}
\end{equation}
where $i$ is the joint index. $z_B$ presents the robot body position in the z-direction. $q_{i}$, $\tau_{i}$ indicate joint angles and joint torques.

The contact force constraint ensures that the vertical component of GRFs remains positive ($f_{z,min}>0$). The friction cone constraint is implemented to prevent the contact point from sliding on the ground. Additionally, the joint velocity and torque constraints ensure that the motor outputs remain within their physical limits, while the joint angle and position constraints are designed to avoid collisions during the jumping motion.

\section{Simulation}\label{section:simulation}
This section focuses on evaluating the performance of the locomotion controller and the optimized jumping trajectory on our quadrupedal wheeled-legged robot in Gazebo simulation environment.
\begin{figure}[htb]
    \centering
    \vspace{-0.0cm}
    \includegraphics[height=0.15\textwidth]{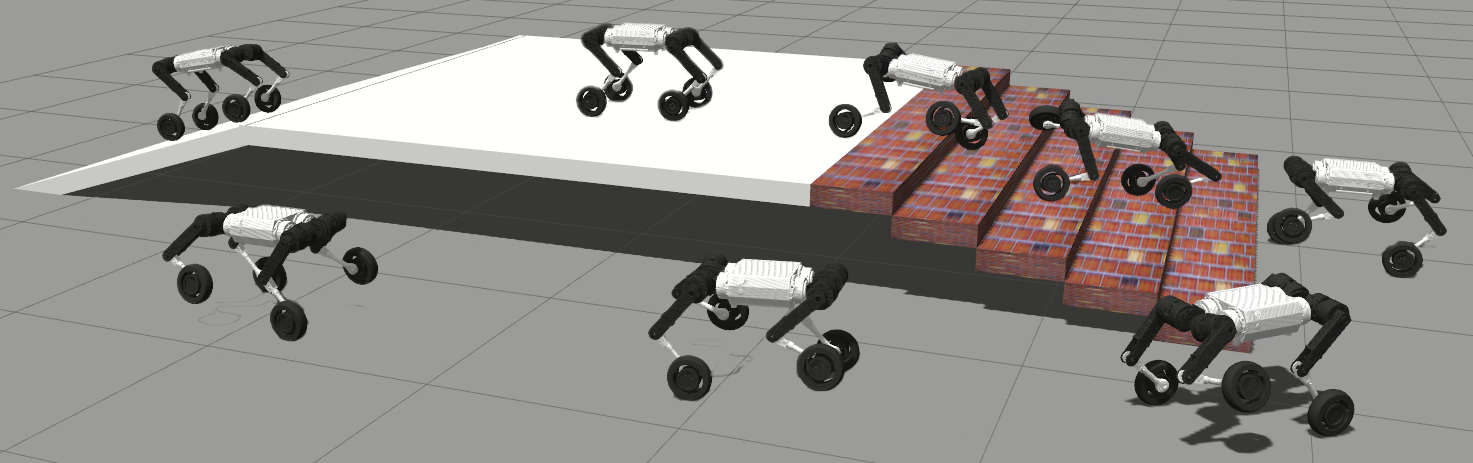}
    \vspace{-0.2cm}
    \caption{The locomotion performance of the robot in the simulation. 
    The one-step height of the stair is 0.1 m, and the slope is 20 degrees.}
    \label{locomotion_sim_result}
    \vspace{-0.3cm}
\end{figure}

\subsection{Locomotion}
In the simulation, the terrain is designed to include a stair with a single-step height of 0.1 m and a slope inclined at 20 degrees. The simulation result (in Fig. \ref{locomotion_sim_result}) demonstrates that the robot performs well in walking through the stairs blindly and rolling down the slope. Furthermore, the integration of wheel motion allows the robot to slide on the ground even when its foot is in contact with the surface. This combination of rolling and trotting gaits provides a significant advantage, enabling the wheeled-legged robot to traverse challenging terrain while maintaining a relatively high moving speed compared to traditional legged robots.

\subsection{Jumping}\label{forward_jump}
Unlike the legged robot, the wheeled-legged robot incorporates rolling motion from its wheels, allowing it to generate initial velocity before executing a jump. In our control framework, this initial velocity is provided by the locomotion controller, while the jumping trajectory optimization module online computes an optimal trajectory based on this initial state. This approach significantly enhances the robot's jumping capabilities, particularly for forward jumps. The detailed sequence of the wheeled-legged robot's jumping motion is depicted in the bottom part of Fig. \ref{control_framework}.

\begin{figure}[htb]
    \centering
    \vspace{-0.2cm}
    \includegraphics[height=0.09\textwidth]{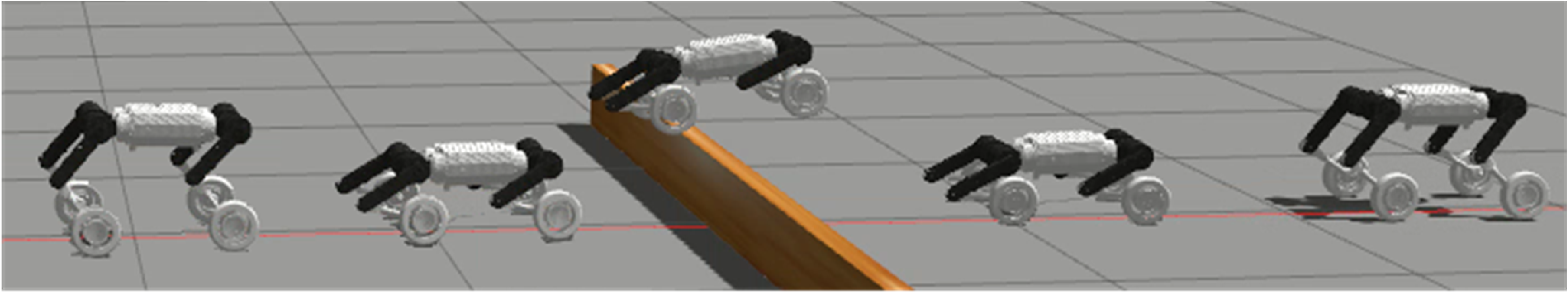}
    \vspace{-0.6cm}
    \caption{The simulation of the forward jumping on the robot. The obstacle is 0.25 m high, and the jumping height of the robot is 0.58 m.}
    \label{forward_jump_sim}
    \vspace{-0.2cm}
\end{figure}

\begin{figure}[htb]
    \centering
    \vspace{-0.4cm}
    \subfigure[The joint velocities]{
    \vspace{-0.5cm}
    \includegraphics[height=0.141\textwidth]{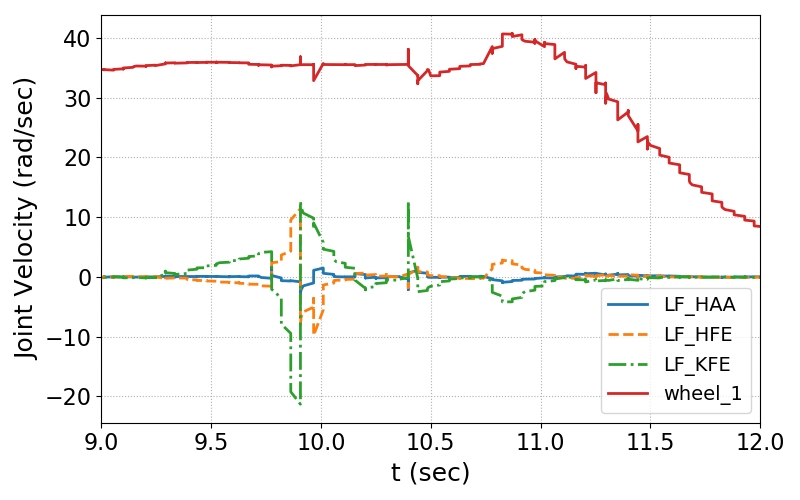}
    \label{jump_vel}
    }
    \subfigure[The joint torques]{
    \vspace{-0.2cm}
    \includegraphics[height=0.141\textwidth]{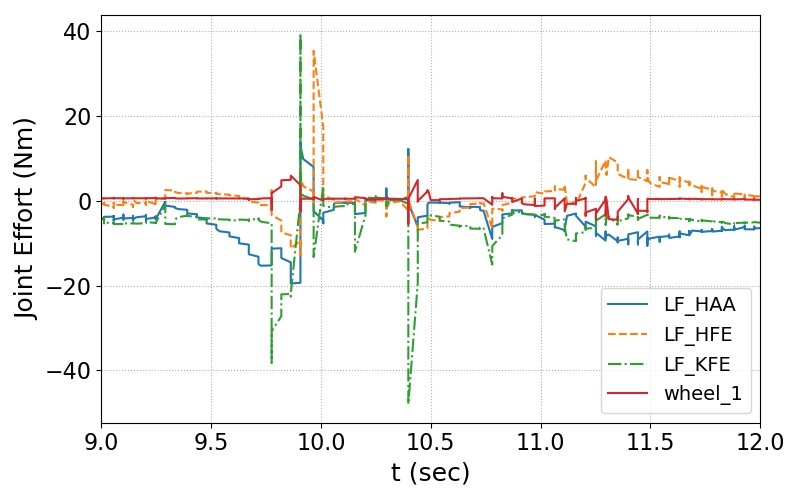}
    \label{jump_torque}
    }
    \vspace{-0.4cm}
    \caption{The joint data of the left front leg during jumping. LF: left front leg; HAA: hip abduction/adduction joint; HFE: hip flexion/extension joint; KFE: knee joint.}
    \label{jump_data}
    \vspace{-0.3cm}
\end{figure}

The first jumping task is aimed at letting the robot jump over an obstacle in an open area. For the jumping optimization problem, the target state at the end of the flight phase $\bm{s}_\Omega(t_3)$ is set to $[0.6, 0.7, 0]$ and $t_1 = t_2$. In the rolling phase (in Fig. \ref{control_framework}), the initial velocity of the robot in the forward direction is achieved by the NMPC (Section \ref{NMP_formulation}). In the simulation, the obstacle is a 0.25 m high wall. Fig. \ref{forward_jump_sim} shows that the robot is able to jump through this obstacle with an initial velocity of 3 m/s, and the highest base position is 0.58 m during jumping. The resultant joint torques and velocities in the jumping are shown in Fig. \ref{jump_data}. It indicates that in such a jumping motion, the knee motor of the robot almost reaches its peak torque (42 Nm), and the velocities of the wheel are at its maximum value (40 rad/s).

\begin{figure}[htb]
    \centering
    \vspace{-0.25cm}
    \subfigure[The simulation snapshots of the backflip]{
    \includegraphics[height=0.135\textwidth]{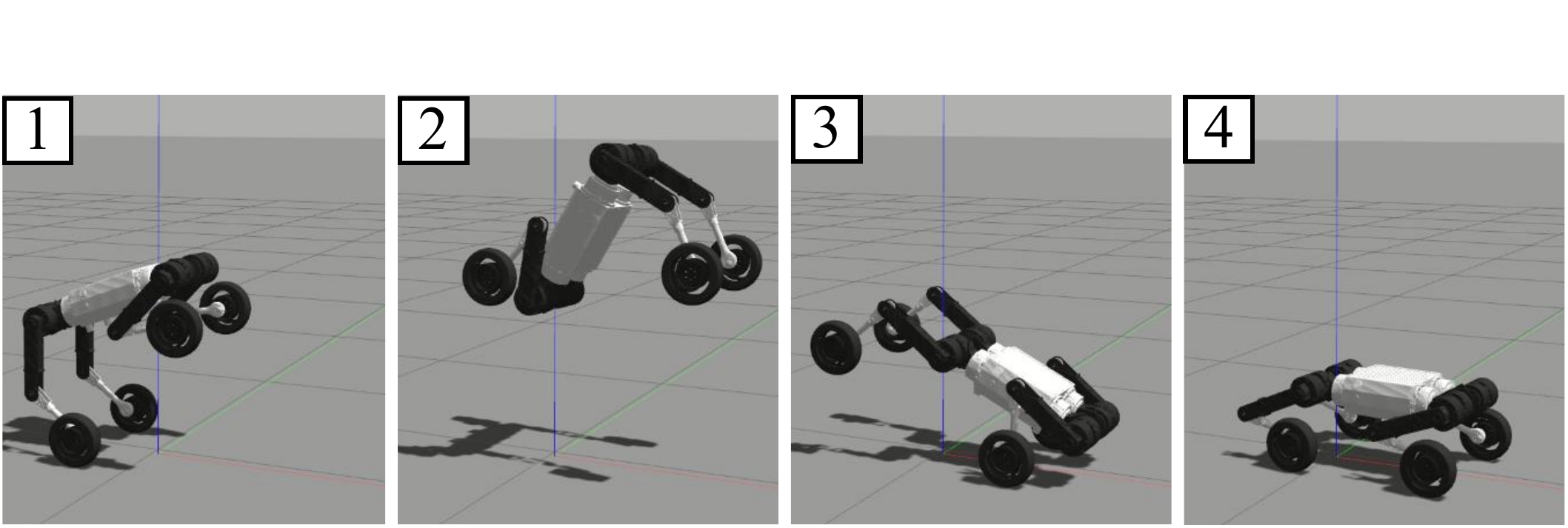}
    }
    \subfigure[The knee joint velocity and torque data of the take-off phase]{
    \includegraphics[width=0.4\textwidth]{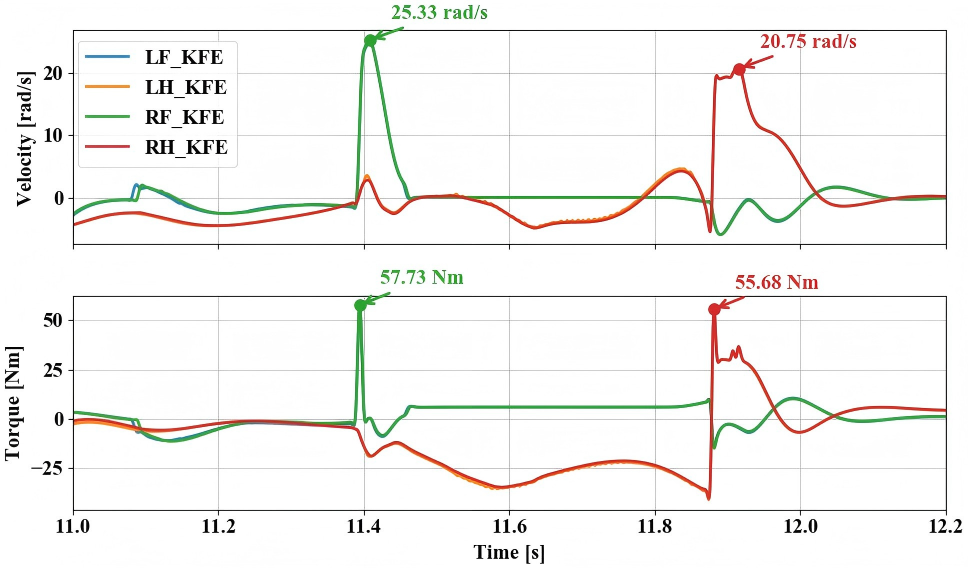}
    \label{data_backflip}
    }
    \vspace{-0.2cm}
    \caption{The backflip simulation. Although the robot achieves the backflip motion successfully by the jumping trajectory optimization, the data indicate that such aggressive motion has already exceeded the hardware limit. }
    \label{black_flip_sim}
    \vspace{-0.45cm}
\end{figure}

The second jumping task is to achieve a backflip motion on the robot. In the simulation, although the robot can do a backflip successfully, as shown in Fig. \ref{black_flip_sim}, by the jumping trajectory optimization (Section \ref{Jumping_opt}) where the target state $\bm{s}_\Omega(t_3)$ is set to $[-0.55, 0.8, 2\pi + \frac{\pi}{2}]$ and $t_1 \neq t_2$, Fig. \ref{data_backflip} denotes that the max torque in the knee joints reaches 57 Nm which has already exceeded the physical limit of the actual motor. Therefore, executing such a backflip motion on our current hardware is not feasible unless we perform weight optimization or use more powerful actuators on the robot.

\section{Experimental Results}
\label{Experiment}
In this section, we validate our control framework in real-world experiments, including the forward jumping and the vertical jumping.

\subsection{Experimental Setup}
The experiments are conducted on our quadrupedal wheeled-legged robot (Fig. \ref{hardware_design}). Each leg of the robot consists of four electric motors: three for joint actuation and one for wheel motion. As a result, the robot has a total of 22 degrees of freedom (DoFs), including 6 from the floating base and 16 from the actuated joints. The wheel module, with a radius of 0.085 m, is mounted at the end of each leg, and each leg link is approximately 0.2 m in length. The total mass of the robot is about 22.5 kg. Each joint actuator, with a gear ratio of 9:1, provides a maximum torque of 42 Nm and a maximum velocity of 40 rad/s.

The NMPC and jumping trajectory optimizer were executed on an onboard Intel NUC BXNUC10I7FNK, equipped with an Intel Core i7 processor and 16 GB of RAM, providing the necessary computational power for real-time control and optimization of the quadrupedal wheeled-legged robot. The locomotion NMPC controller operates at 50 Hz, with a prediction horizon of 1 second and a discrete time step of 0.015 seconds. The jumping optimization is capable of finding the solution within 1 second.
\begin{figure}[htb]
    \centering
    \vspace{-0.1cm}
    \includegraphics[height=0.21\textwidth]{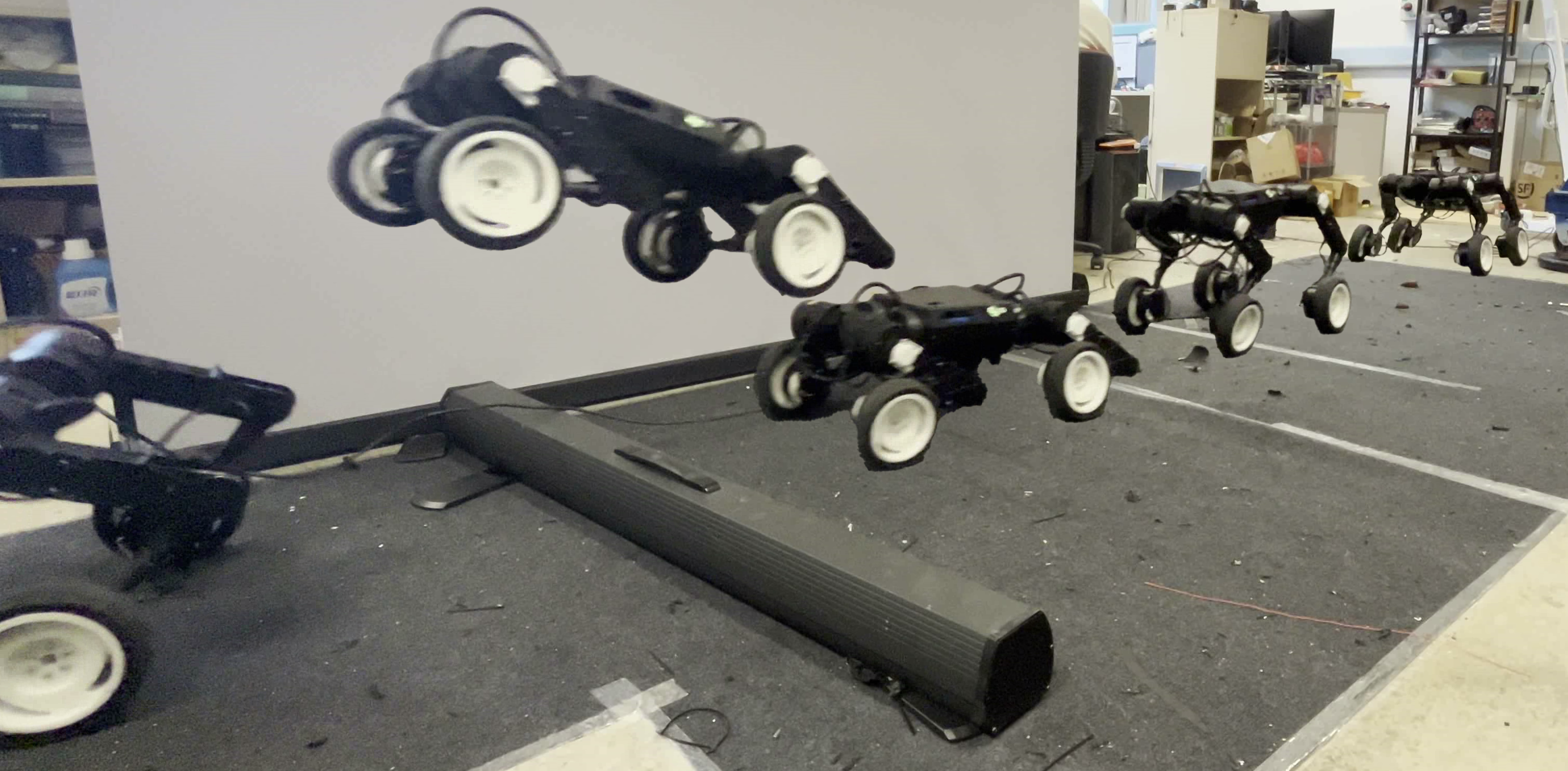}
    \vspace{-0.1cm}
    \caption{The snapshot of the forward jumping experiment. }
    \label{forward_jump_front_view}
    \vspace{-0.5cm}
\end{figure}

\subsection{Results}
The first hardware experiment is for evaluating the forward jumping performance of the robot. The desired state of the robot $\bm{s}_\Omega(t_3)$ is defined as $[0.6, 0.7, 0]$, consistent with the parameters used in the simulation for the forward jumping (Section \ref{forward_jump}). The experiment proves that the robot can achieve a forward jump over a 0.12 m obstacle successfully although the robot slightly contacts the obstacle due to the fact that the operator is hard to determine the jumping timing, as shown in Fig. \ref{foward_jump_1} and Fig. \ref{forward_jump_front_view}. With the advantage of the wheel motion, the robot is able to jump farther compared with legged robots. The corresponding data of the joint torque and velocity of the front left leg is illustrated in Fig. \ref{forward_tau} and Fig. \ref{forward_vel}. Fig. \ref{forward_vel} shows that the wheel speed is about 20 rad/s before the forward jump.

In the second experiment, the target for the jumping trajectory optimization is set to achieve a vertical jump height of 0.7 m. During the real-world experiment, the robot reaches a vertical jump height of approximately 0.5 m, as illustrated in Fig. \ref{vertical_jump_1}. The corresponding joint torque and velocity profiles during this motion are displayed in Fig. \ref{vertical_tau} and Fig. \ref{vertical_vel}, respectively.

The obtained results validate the effectiveness of the proposed control framework in optimizing the jumping motion for the quadrupedal wheeled-legged robot. Notably, the robot has a total mass of 22.5 kg, which is relatively heavy for a robot of its size. This is primarily due to the weight of the wheel modules, with each module weighing 1.5 kg. Despite the significant mass, the robot demonstrates impressive jumping performance in the experiments. As shown in Fig. \ref{joint_data_exp}, the maximum torque of the knee joint reaches 30 Nm, and the maximum joint velocity achieves approximately 14 rad/s during both forward and vertical jumping tasks.
\begin{center}
\vspace{-0.4cm}
\begin{figure}[htb]
\centering
\subfigure[Torque (forward jumping)]{
    \vspace{-0.1cm}
    \includegraphics[height=0.165\textwidth]{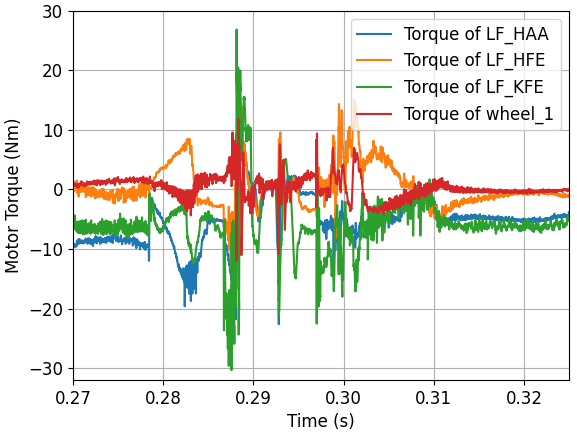}
    \label{forward_tau}
    }
\subfigure[Velocity (forward jumping)]{
    \vspace{-0.3cm}
    \includegraphics[height=0.163\textwidth]{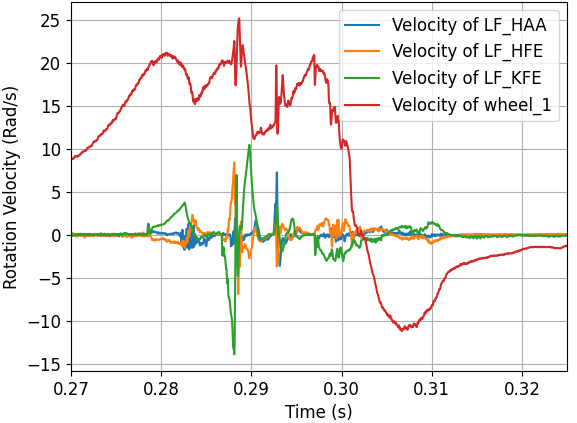}
    \label{forward_vel}
    }
\subfigure[Torque (vertical jumping)]{
    \vspace{-0.3cm}
    \includegraphics[height=0.162\textwidth]{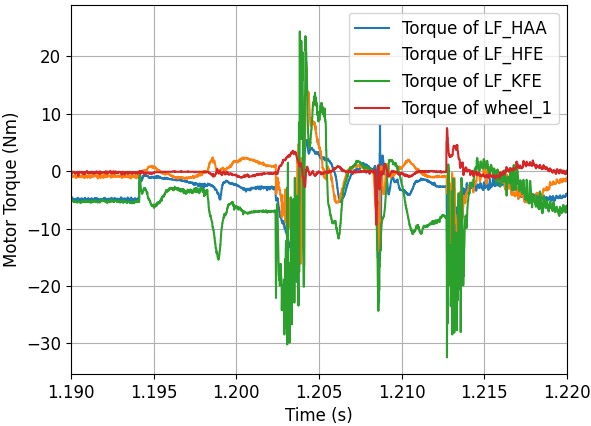}
    \label{vertical_tau}
    }
\subfigure[Velocity (vertical jumping)]{
    \vspace{-0.3cm}
    \includegraphics[height=0.161\textwidth]{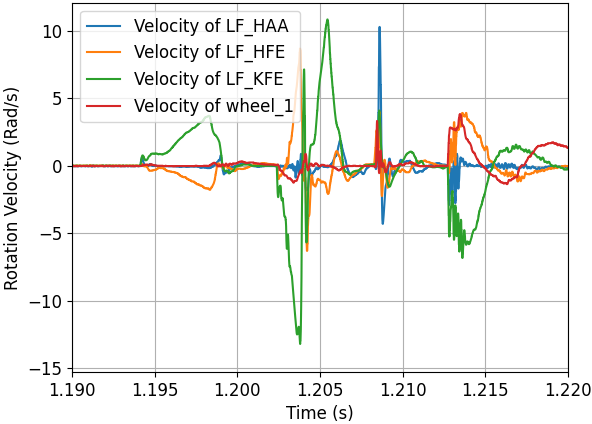}
    \label{vertical_vel}
    }
    \vspace{-0.4cm}
\caption{The joint data of the robot's left front leg during the forward jumping and vertical jumping experiments.}
\label{joint_data_exp}
\end{figure}
\vspace{-0.3cm}
\end{center}

\section{CONCLUSIONS}
\label{Conclusion}
This paper presents the design of a mini-sized quadrupedal wheeled-legged robot and introduces a motion control framework for enabling the robot's jumping. By NMPC for locomotion and trajectory optimization via the DE algorithm, the proposed approach successfully enhances the robot’s jumping capabilities. Simulations and hardware experiments on our robot confirm the effectiveness of the method, showcasing backflip, forward jumps, and vertical jumps. However, there is a gap between simulations and experiments, which is due to the trade-off relation between motor torque and velocity \cite{motor_limit} in the real hardware. The results highlight the advantages of integrating wheel motion into jumping strategies, allowing for greater jump distances and improved obstacle traversal. Future work will explore optimizing hardware design to further improve jumping performance and extending the control framework to handle more complex terrains and dynamic environments.

\addtolength{\textheight}{0cm}   









\end{document}